\documentclass[10pt,twocolumn,letterpaper]{article}

\usepackage{cvpr}              

\usepackage[dvipsnames]{xcolor}
\usepackage{adjustbox}
\usepackage{makecell}
\usepackage{algorithmic}
\usepackage{graphicx}
\usepackage{textcomp}
\usepackage{comment}
\usepackage{soul,xcolor}

\definecolor{cvprblue}{rgb}{0.21,0.49,0.74}
\usepackage[pagebackref,breaklinks,colorlinks,citecolor=cvprblue]{hyperref}
\usepackage[accsupp]{axessibility}

\def\paperID{6} 
\def\confName{CVPR}
\def\confYear{2024}

\title{Enhancing Visual Question Answering\\through Question-Driven Image Captions as Prompts}

\author{Övgü Özdemir, Erdem Akagündüz\\
Department of Modeling and Simulation,\\
Graduate School of Informatics,\\
Middle East Technical University, Türkiye\\
{\tt\small{\{ovgu.ozdemir,akaerdem\}@metu.edu.tr}}
}

\begin{document}
\maketitle
\begin{abstract}
Visual question answering (VQA) is known as an AI-complete task as it requires understanding, reasoning, and inferring about the vision and the language content. 
Over the past few years, numerous neural architectures have been suggested for the VQA problem. 
However, achieving success in zero-shot VQA remains a challenge due to its requirement for advanced generalization and reasoning skills.
This study explores the impact of incorporating image captioning as an intermediary process within the VQA pipeline. 
Specifically, we explore the efficacy of utilizing image captions instead of images and leveraging large language models (LLMs) to establish a zero-shot setting.
Since image captioning is the most crucial step in this process, we compare the impact of state-of-the-art image captioning models on VQA performance across various question types in terms of structure and semantics.
We propose a straightforward and efficient question-driven image captioning approach within this pipeline to transfer contextual information into the question-answering (QA) model. 
This method involves extracting keywords from the question, generating a caption for each image-question pair using the keywords, and incorporating the question-driven caption into the LLM prompt. We evaluate the efficacy of using general-purpose and question-driven image captions in the VQA pipeline.
Our study highlights the potential of employing image captions and harnessing the capabilities of LLMs to achieve competitive performance on GQA under the zero-shot setting. Our code is available at \url{https://github.com/ovguyo/captions-in-VQA}.
\end{abstract}    
\section{Introduction}
\label{sec:intro}

Visual Question Answering (VQA) is a complex multimodal task that demands a high-level understanding of several aspects, such as object and attribute identification, object localization, comprehension of the relationship between the image and the question, and reasoning about the context and the scene. The common steps of a typical VQA model involve generating embeddings of the image and the question using encoders for each, combining the image and question embeddings with a fusing module, and generating answers using a text generator or a classifier. For a general overview of the VQA techniques, the reader may refer to  \cite{srivastava2021visual, sharma2021survey}.

The inherent multimodal nature of the VQA problem is the primary factor contributing to its complexity.
Combining different types of information, such as text and images, makes the model's training more complex, as the model must understand and utilize the connections and interactions between these different modalities. 
Several studies \cite{kamath2021mdetr, santoro2017simple, perez2018film, ben2017mutan, hudson2018compositional} propose an approach to tackle multimodality for the VQA problem. 
However, these methods indicate limitations in their capacity to adapt to new tasks, particularly in zero-shot settings. 

Recent advances in high-capacity large language models (LLMs) \cite{brown2020language, achiam2023gpt, team2023gemini} have marked a dramatic milestone in the domain. LLMs are predominantly trained with millions (or billions) of parameters and utilized for processing textual data. LLMs show outstanding performance in a variety of natural language tasks.
The ongoing research challenge lies in extending the capabilities of LLMs to the intersection of different modalities, \eg, textual and visual data. Recently, GPT-4 \cite{achiam2023gpt} and Gemini \cite{team2023gemini} stand out as remarkable examples of multimodal LLMs, adept at successfully processing textual and visual modalities for various downstream tasks, including VQA. Several alternative approaches \cite{li2022blip, li2023blip, liu2024visual, alayrac2022flamingo, dai2023instructblip} have also been proposed in the realm of large-scale vision-language integration. 
The challenge in multimodal training lies in the extensive computational and data costs required to align the representation spaces of vision and language.

Some recent studies \cite{guo2023images, tiong2022plug, yang2022empirical} delve into the potential of utilizing image captions with unimodal LLMs in the zero-shot VQA setting. 
Our study differs from these studies in the following aspects. Firstly, we focus on examining the representation capacity of image captions from various vision-language models on the VQA performance. Second, our study investigates whether image captions can be informative for specific types of questions by evaluating the results in structurally and semantically different questions. Within this scope, we also evaluate the influence of feeding LLMs with general-purpose and question-driven captions, and only the most relevant sentence in the caption during the QA stage.

Numerous VQA datasets are available in the literature, including CLEVR \cite{johnson2017clevr}, VQA \cite{antol2015vqa}, VQA 2.0 \cite{goyal2017making}, OK-VQA \cite{marino2019ok}, GQA \cite{hudson2019gqa}. Among these sets, although each serves various purposes effectively, GQA stands out for its emphasis on testing compositional and grounded reasoning abilities and its relatively diverse Q/A set. In this study, we conduct our experiments on the GQA dataset and focus on measuring performance on semantically and structurally different questions.

We structure the VQA task into two fundamental components: image captioning and question-answering. The goal is to leverage the respective strengths of these tasks, aiming for a more thorough comprehension of both the visual content and the corresponding questions. 
We carry out experiments with state-of-the-art vision-language models, including CogVLM \cite{wang2023cogvlm}, BLIP-2 \cite{li2023blip}, and FuseCap \cite{rotstein2024fusecap} to comprehend their scene representation capacity in the VQA pipeline.

We outline our contributions as follows:

\begin{itemize}
    \item We evaluate the image captioning performance of various vision-language models incorporating them with LLMs for zero-shot VQA, analyzing their effectiveness across various question types.
    \item We propose a straightforward question-driven captioning approach to better transfer the context into LLMs for question-answering.
\end{itemize}
The rest of the paper is organized as follows. Section \ref{sec:related} reviews related works. Section \ref{sec:methodology} mentions the components of the proposed pipeline. In Section \ref{sec:experiments}, we present the experiments designed for our study. Section \ref{sec:results} discusses evaluation results. Section \ref{sec:conclusion} outlines the conclusions drawn from our findings and discusses potential avenues for future research.

\section{Related Literature}
\label{sec:related}

\subsection{Large Language Models}
\label{sec:largelanguage}

LLMs \cite{brown2020language, chatgpt, touvron2023llama} trained on extensively rich web-scale corpus usually employ autoregressive methods to generate target tokens. LLMs demonstrate remarkable proficiency in processing and generating text with human-like characteristics. This attribute renders them suitable instruments for various language-related tasks, including question-answering, text generation, machine translation, etc.
Expanding the scope of LLMs to include additional modalities results in the creation of multimodal LLMs \cite{li2023blip, achiam2023gpt, wang2023cogvlm, team2023gemini, liu2024visual}, which boosts the performance for many downstream tasks including image captioning, visual question answering, text-to-image synthesis.  

\subsection{Visual Question Answering}
\label{sec:vqa}

The main challenge of the VQA domain comes from bridging the gap between visual understanding and natural language. 
Numerous studies have been proposed to tackle questions related to visual content. 
Relation Networks \cite{santoro2017simple} involves employing a compact and straightforward neural network module that takes pairs of features as input and generates a score indicative of the relationship between these feature pairs.
LXMERT \cite{tan2019lxmert} is a large-scale transformer model that fuses textual and visual representations with a cross-modality encoder.
MDETR \cite{kamath2021mdetr} is an end-to-end modulated detector which is an improved version of the object detection model DETR \cite{carion2020end} by adding the capability of processing free-form texts. 
Alternatively, neuro-symbolic approaches in VQA have gained attention to enhance model interpretability. A neuro-symbolic approach in VQA combines two main parts: neural network modules for handling images and text modalities, and a symbolic reasoning module for managing logic and knowledge representation. 
NS-VQA \cite{yi2018neural} and NS-CL \cite{mao2019neuro} use neural networks for scene parsing and dividing questions into program instructions, and propose a symbolic module executing the program instructions on the scene representation. 
An alternative hybrid approach, ProTo \cite{zhao2021proto}, proposes program-guided transformers that use semantic and structural information of the programs being parsed from the questions by a sequence-to-sequence model. 
A recent approach, namely VisProg \cite{gupta2023visual}, generates program instructions from questions using LLMs and employs instructions on images benefiting from different modules for object detection, visual question answering, image classification, and segmentation. Recent large-scale multimodal approaches used for VQA are mentioned in Section \ref{sec:intro} and Section \ref{sec:largelanguage}.

\subsection{Image Captioning}
\label{sec:image-captioning}

Image captioning aims to produce a caption describing visual content in natural language. 
Conventional approaches in image captioning are based on attention and encoder-decoder structure \cite{huang2019attention, xu2015show, herdade2019image}.
A typical image captioning model consists of an encoder for gathering visual cues and a textual decoder to produce the final caption. 
Like VQA, this requires bridging the gap between visual and natural language understanding.
Recently, large-scale multimodal models \cite{mokady2021clipcap, li2022blip, gupta2023visual, wang2023cogvlm, achiam2023gpt, team2023gemini, li2023blip} have resulted in notable enhancements in performance and demonstrated adaptability to various downstream applications, including image captioning.

\subsection{Question Answering}
\label{sec:qa}

Question-answering (QA) models aim to provide contextually appropriate responses based on a document or text, often requiring an understanding of linguistic rules, syntax, and contextual nuances.
Recent models in QA leverage transformer architectures and large-scale pre-training on diverse datasets \cite{devlin2018bert, liu2019roberta, radford2019language, brown2020language}.

\begin{figure*}[th!]
\centering
    \caption{VQA pipeline exploiting general and the proposed question-driven (QD) image captioning as an intermediate step.}
    \label{fig:pipeline1}
    \includegraphics[width=0.75\paperwidth]{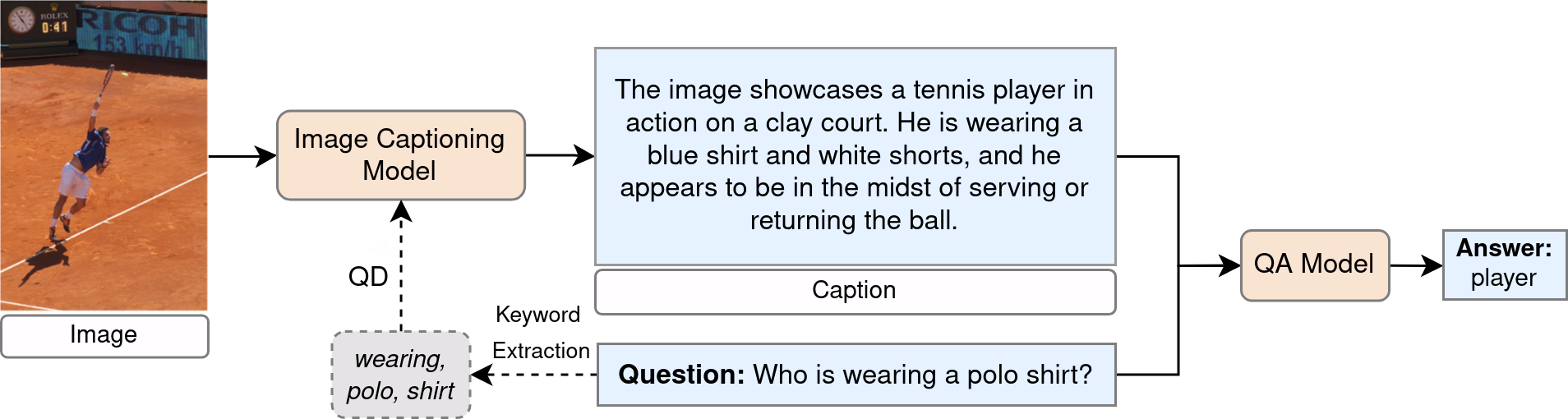}
\end{figure*}

\section{Methodology}
\label{sec:methodology}

\subsection{Caption Generation}
The primary and most crucial element in the suggested pipeline is the creation of image captions with high visual representation capability. Image captions provide a summarized version of the visual content, and specific visual details may be lost, which could affect the VQA performance. We survey image captioning models, selecting ones that provide more detailed captions while taking into account our computational resource limitations. Consequently, we evaluate several zero-shot vision-language models, including CogVLM \cite{wang2023cogvlm}, FuseCap \cite{rotstein2024fusecap}, and BLIP-2 OPT$_{2.7b}$ \cite{li2023blip} by integrating them into the VQA pipeline. We employ both the chat and visual grounding variants of CogVLM, considering their potential performance impacts across different question types.
VQA performance is assessed across various image captions according to structurally and semantically different question categories. More details about question categories are given in Section \ref{sub:dataset}.

Two approaches are utilized in this paper to generate captions. First, each image is captioned without considering the questions associated with it, which we refer to as  ``general-purpose captioning'' throughout the paper. However, general-purpose captions are designed to provide a broad description of the visual content, and they may lack the precision needed to address detailed and specific queries. Therefore, in our second approach, we create image captions for each image-question pair, a process we refer to as  ``question-driven image captioning''. For this purpose, KeyBERT \cite{grootendorst2020keybert} is employed to extract keywords from the questions. KeyBERT utilizes BERT-embeddings along with a basic cosine similarity measure to identify the most representative words that encapsulate the content of the entire text. Extracted keywords are fed into the image captioning model along with the corresponding image, as illustrated in Figure \ref{fig:pipeline1}. 

We also investigate whether less relevant portions of an image caption could potentially introduce confusion or result in inaccurate answers for the QA model/LLM. Hence, in our analysis, we experiment with keeping only the most relevant sentence of the image caption and providing it to the LLM during the QA step. To achieve this, we utilize Sentence-BERT \cite{reimers2019sentence}, specifically employing the MiniLM-L6 model\footnote{https://huggingface.co/sentence-transformers/multi-qa-MiniLM-L6-cos-v1}, to extract the most relevant sentence from the image caption based on the given question.

\subsection{Question Answering}
As in the pipeline shown in Figure \ref{fig:pipeline1}, the QA model takes the image caption and the question as input, leveraging information from the image caption to generate an answer.  
During the QA step, we utilize GPT-3.5, recognized for its high zero-shot performance in QA benchmarks \cite{wang2024evaluating}. 
Despite the superior performance of the more recent LLM, GPT-4, across various natural language tasks including QA, we choose not to use GPT-4 in our experiments to keep our pipeline cost-effective. In future works, the integration of higher-performing LLMs with the pipeline could be explored.

We derive answers with an open-ended generation, specifically using GPT-3.5-turbo API provided by OpenAI. The answer size is restricted to a maximum of two words, aligning with the answer size distribution in the GQA dataset. Optimal prompts are given in the Section \ref{sub:prompt-details}.

\section{Experimental Setup}
\label{sec:experiments}

\subsection{Dataset} 
\label{sub:dataset}
We conduct experiments on the GQA \cite{hudson2019gqa} dataset, specifically the balanced version of the test-dev subset, comprising $12,578$ questions. Each image in the dataset is linked to multiple questions, and the overall number of images included is $398$. This subset contains a diverse distribution of questions across various categories, with a primary focus on categorization based on structure and semantics. The structural type is determined by the final operation in the functional program of the question, encompassing categories such as \emph{verify, query, choose, logical,} and \emph{compare}. The semantic type specifies the primary focus of the question and includes categories like  \emph{object, attribute, category, relation,} and \emph{global}.
Table \ref{tab:question-types} presents an overview of question types, corresponding descriptions, and the respective number of questions in the GQA test-dev \cite{hudson2019gqa}.

\begin{table*}[h]
\caption{\label{tab:question-types} Overview of the question types}
\begin{center}
\scalebox{0.85}{
\begin{tabular}{ |c c c c| } 
 \hline
    Question type & Description & Example & No. samples \\
    \hline
    verify & yes/no questions & \emph{Does the device under the picture frame look black?} & 2252 \\
    query & open questions & \emph{Which kind of vehicle is waiting for the traffic light?} & 6805 \\
    choose & choosing from alternatives & \emph{What color is the hair, gray or red?} & 1128 \\
    logical & logical inference & \emph{Are the flags triangular and red?} & 1803 \\
    compare & comparison of objects & \emph{Which is larger, the pasture or the horse?} & 589 \\
    \hline
    object & existence questions & \emph{Are there both a horse and a fence in the image?} & 778 \\
    attribute & object properties/position  & \emph{On which side of the picture are the pens?} & 5185 \\
    category & object identification & \emph{What kind of clothing is yellow?} & 1149 \\
    relation & relations with objects/subjects &  \emph{Is the toaster to the right of a refrigerator?} & 5308 \\
    global & overall properties & \emph{Is it an outdoors scene?} & 157 \\
    
  \hline
\end{tabular}}
\end{center}
\end{table*}

\subsection{Competing VQA Methods}
To evaluate zero-shot VQA performance, we use the chat variation of CogVLM\footnote{https://huggingface.co/THUDM/cogvlm-chat-hf} and BLIP-2 FlanT5$_{XL}$\footnote{https://huggingface.co/Salesforce/blip2-flan-t5-xl}. CogVLM is an open-sourced pre-trained vision-language model with 10B visual and 7B language parameters. CogVLM outperforms many vision-language models, \eg, InstructBLIP \cite{dai2023instructblip} and LlaVA-1.5 \cite{liu2023improved}, in VQA benchmarks. Due to our resource constraints with $16$ GB VRAM, we apply $4$-bit quantization to CogVLM. BLIP-2 FlanT5$_{XL}$ with 4.1B parameters also indicate high performance surpassing BLIP-2 OPT$_{6.7B}$ and Flamingo \cite{alayrac2022flamingo} in VQA benchmarks. We employ BLIP-2 FlanT5$_{XL}$ with F16 precision.

\subsection{Image Captioning Methods}
We examine the VQA performance attributed to semantic and structural question types mentioned in Section \ref{sub:dataset}. Image captions are obtained through the visual grounding\footnote{https://huggingface.co/THUDM/cogvlm-grounding-generalist-hf} and chat\footnote{https://huggingface.co/THUDM/cogvlm-chat-hf} variations of the CogVLM, FuseCap\footnote{https://github.com/RotsteinNoam/FuseCap}, and BLIP-2 OPT$_{2.7b}$\footnote{https://huggingface.co/Salesforce/blip2-opt-2.7b} models. When determining the image captioning method, we pay attention to both its alignment with our resource capacity and its high performance in image captioning benchmarks. We employ 4-bit quantization to CogVLM and use F16 precision for BLIP-2 OPT$_{2.7b}$. 

\subsection{Evaluation}
Before the evaluation, GPT-3.5 predictions undergo post-processing, which involves the removal of punctuation. During the evaluation process, we employ the accuracy metric, calculated as the ratio of correctly predicted answers to the total number of answers. Given that answers are derived through open-ended generation using LLMs and might include variations, we do not seek an exact match between the prediction and the ground truth. Instead, we evaluate semantic similarity using cosine similarity in a vector space with the threshold $0.70$. If two strings are closely aligned in meaning, the prediction is accepted as correct; for example, accepting $couch$ as correct for the label $sofa$. We determine the similarity threshold through manual observation of the results. At lower thresholds, we observe that predictions incorporating words related to each other, yet lacking identical meanings, are also considered correct. For instance, the similarity value between the words \emph{blue} and \emph{brown} is found to be $0.67$. We additionally assess performance across higher cosine similarity thresholds, \eg $0.8$ and $0.9$, and for exact matching (EM).

\begin{table*}[hbt!]
\caption{\label{tab:image-captioning} Comparison of the performances of different image captioning methods in the context of VQA on GQA test-dev. Image captioning methods are employed with GPT-3.5 as the question-answering (QA) method. Two variants of CogVLM, namely visual grounding (CogVLM-V) and chat model (CogVLM-C), are utilized for image captioning. QD and SB refer to question-driven and sentence-based captions, respectively. The answers with a cosine similarity of $0.7$ or higher have been considered correct with the label. Accuracy values are compared with the performance of zero-shot VQA models based on various question categories. }
\begin{center}
\scalebox{0.95}{
\begin{tabular}{ |c|c|c|c|c|c|c||c|c| } 
 \hline
    \thead{Question \\ type} & \thead{CogVLM-C\\ Cap.\\+ GPT-3.5 QA} & \thead{CogVLM-V\\ Cap.\\+ GPT-3.5 QA} & \thead{CogVLM-C\\ QD Cap.\\+ GPT-3.5 QA} & \thead{CogVLM-C\\ SB Cap.\\+ GPT-3.5 QA} & \thead{FuseCap Cap.\\+ GPT-3.5 QA} & \thead{BLIP-2 Cap.\\+ GPT-3.5 QA} & \thead{CogVLM VQA} & \thead{BLIP-2 VQA} \\
  \hline
 verify & 63.01 & 58.53 & \textbf{66.83} & 61.06 & 53.60 & 55.82 & \textbf{83.04} & 56.48\\ 
 query & 36.91 & 31.08 & \textbf{38.34} & 31.51 & 29.61 & 31.87 & \textbf{54.11} & 41.31 \\ 
 choose  & 65.25 & 60.90 & \textbf{65.51} & 60.90 & 58.07 & 60.82 & \textbf{87.32} & 56.91 \\ 
 logical & 59.51 & \textbf{60.29} & 59.07 & 58.46 & 57.07 & 56.07 & \textbf{77.54} & 54.24 \\ 
 compare & 51.78 & 51.95 & 51.95 & 49.07 &\textbf{54.50} & 48.22 & \textbf{62.65} & 46.52 \\ 
  \hline
 object & 61.95 & \textbf{63.24} & 59.13 &  58.87 & 59.38 & 58.35 & \textbf{84.45} & 57.07 \\ 
 attribute & 51.75 & 46.42 & \textbf{54.62} & 50.80 & 45.11 & 46.63 & \textbf{70.45} & 49.33 \\ 
 category & 47.35 & 44.21 & \textbf{50.39} & 42.56 & 43.52 & 42.47 & \textbf{63.19} & 53.35 \\ 
 relation & 42.56 & 38.32 & \textbf{42.97} & 35.76 & 34.98 & 37.23 & \textbf{59.91} & 43.31 \\ 
 global & \textbf{49.04} & 45.86 & 45.86 & 44.59 & 43.95 & 45.22 & \textbf{56.05} & 40.13 \\ 
  \hline
 total & 48.06 & 43.83 & \textbf{49.50} & 44.12 & 41.58 & 42.99 & \textbf{66.02} & 47.52 \\  
 \hline

\end{tabular}}
\label{tab1}
\end{center}
\end{table*}

\subsection{Prompt Details} 
\label{sub:prompt-details}
A brief prompt, `Describe the scene in this image' is supplied to the image captioning model to create general-purpose image captions. To create question-driven captions, `Consider the keywords: \texttt{[keywords]}' is added to the prompt. In the QA stage, the LLM prompt involves `Answer the question in a maximum of two words based on the text. Consider the type of question in your answer. For example, if it is a yes/no question, the answer should be yes or no. Text: \texttt{[text]}, Question: \texttt{[question]}'. We notice a positive impact on the results when we include an instruction in the prompt to consider the question type. In the decoding step for the answer generation, we set $temperature$ as $0.2$, $top_p$ as 1, and specify $frequency\_penalty$ and $presence\_penalty$ as 0.

\section{Results}
\label{sec:results}

\begin{figure*}
\begin{subfigure}{.24\textwidth}
  \includegraphics[trim= 0 210 0 0,clip,width=.95\linewidth]{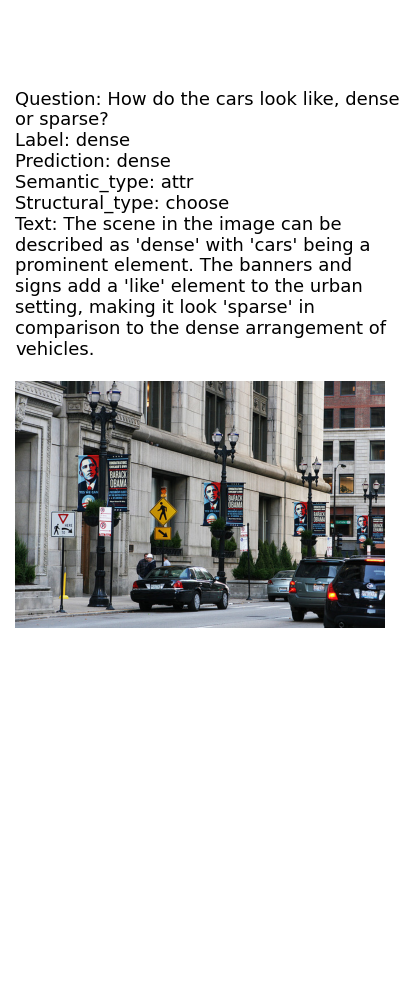}
  \label{fig:sfig1}
\end{subfigure}%
\begin{subfigure}{.24\textwidth}
  \includegraphics[trim= 0 210 0 0,clip,width=.95\linewidth]{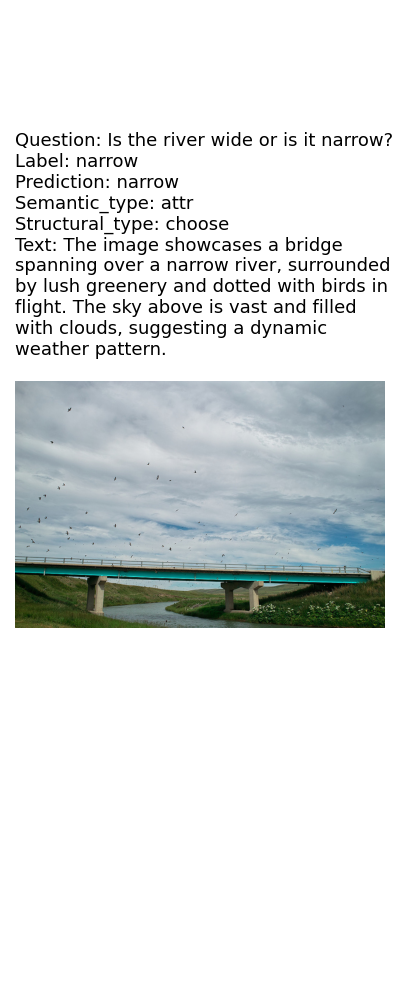}
  \label{fig:sfig2}
\end{subfigure}
\begin{subfigure}{.24\textwidth}
  \includegraphics[trim= 0 210 0 0,clip,width=.95\linewidth]{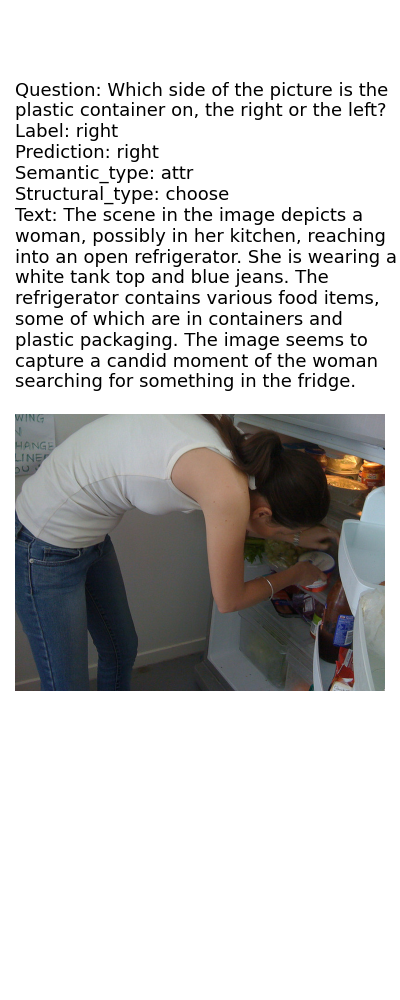}
  \label{fig:sfig3}
\end{subfigure}
\begin{subfigure}{.24\textwidth}
  \centering
  \includegraphics[trim= 0 210 0 0,clip,width=.95\linewidth]{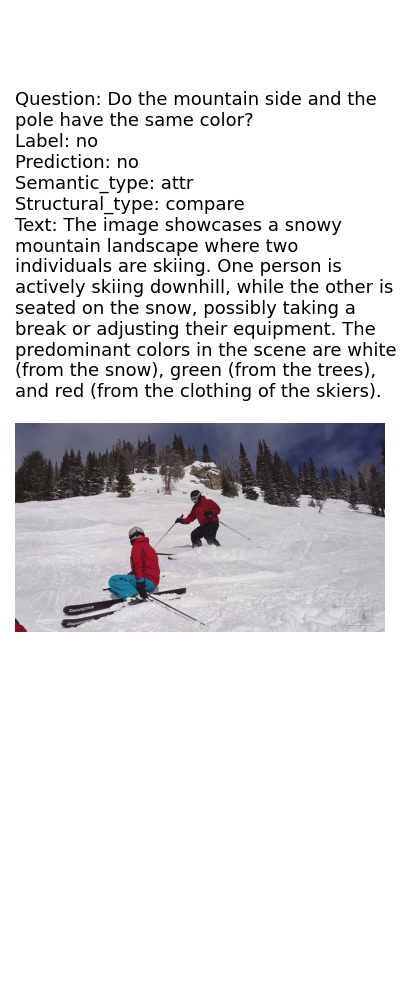}
  \label{fig:sfig4}
\end{subfigure}
\begin{subfigure}{.24\textwidth}
  \centering
  \includegraphics[trim= 0 210 0 0,width=.95\linewidth]{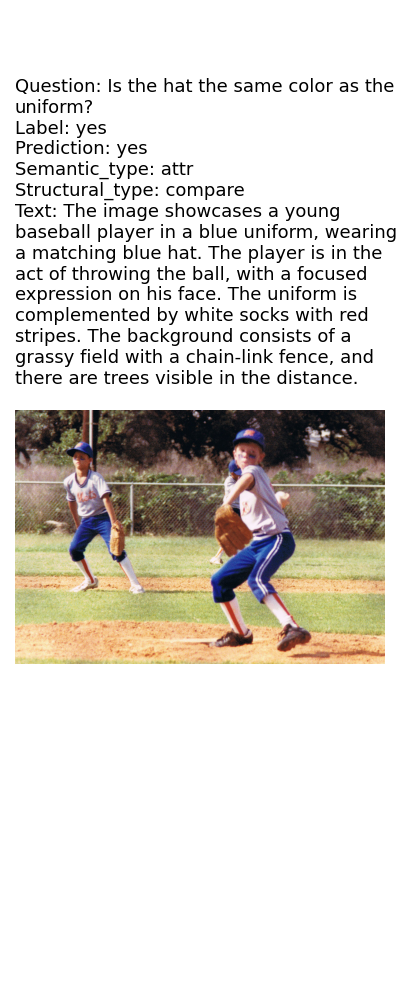}
  \label{fig:sfig5}
\end{subfigure}
\begin{subfigure}{.24\textwidth}
  \centering
  \includegraphics[trim= 0 210 0 0,width=.95\linewidth]{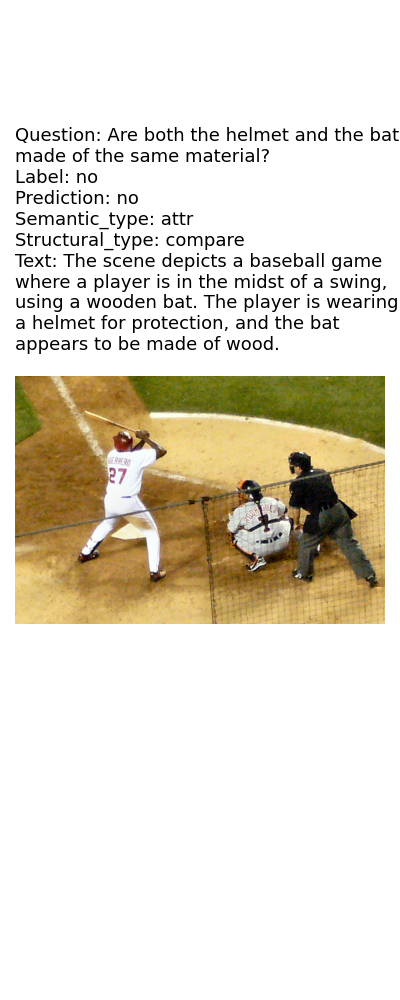}
  \label{fig:sfig6}
\end{subfigure}
\begin{subfigure}{.24\textwidth}
  \centering
  \includegraphics[trim= 0 210 0 0,width=.95\linewidth]{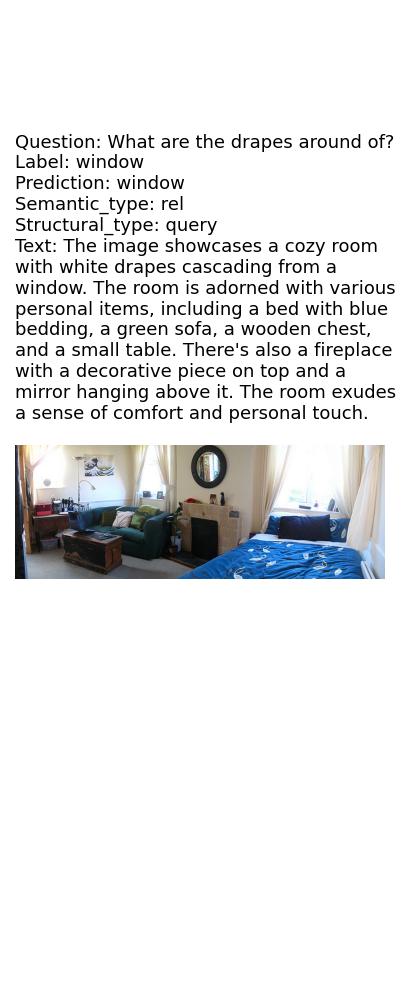}
  \label{fig:sfig7}
\end{subfigure}
\begin{subfigure}{.24\textwidth}
  \centering
  \includegraphics[trim= 0 210 0 0,width=.95\linewidth]{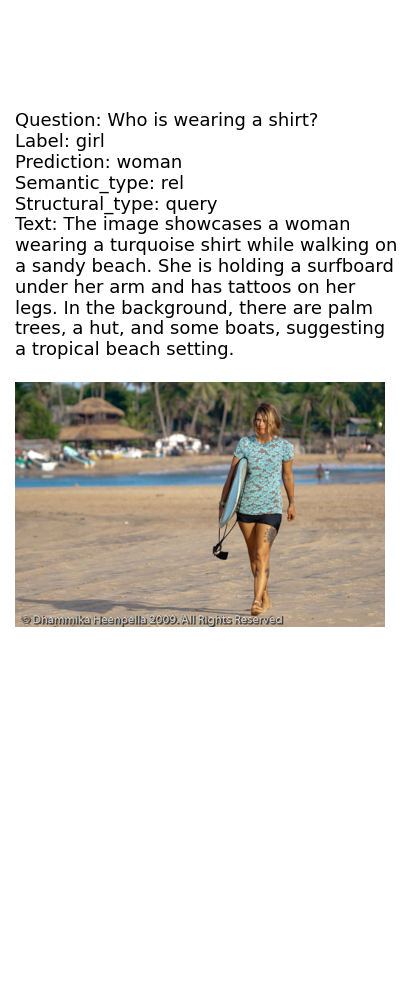}
  \label{fig:sfig8}
\end{subfigure}
\begin{subfigure}{.24\textwidth}
  \centering
  \includegraphics[trim= 0 210 0 0,width=.95\linewidth]{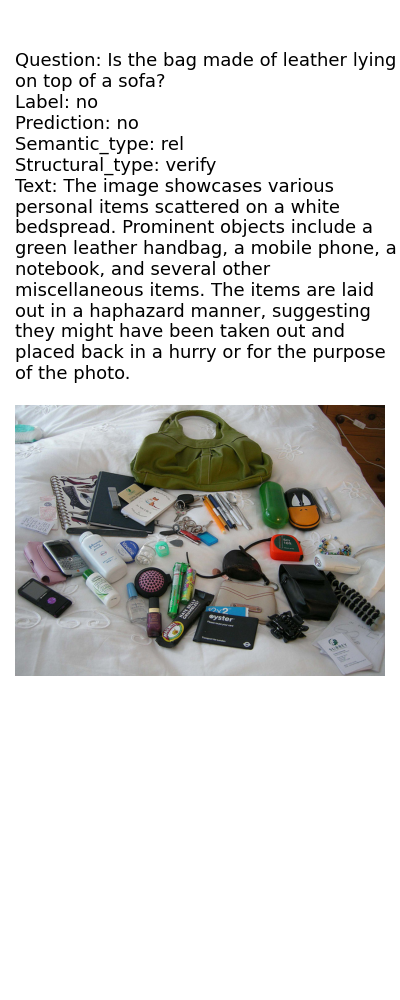}
  \label{fig:sfig9}
\end{subfigure}
\begin{subfigure}{.24\textwidth}
  \centering
  \includegraphics[trim= 0 210 0 0,width=.95\linewidth]{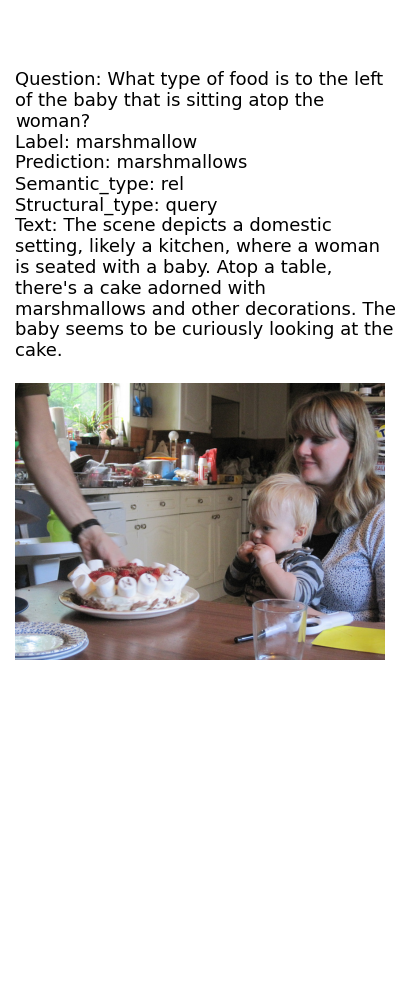}
  \label{fig:sfig10}
\end{subfigure}
\begin{subfigure}{.24\textwidth}
  \centering
  \includegraphics[trim= 0 210 0 0,width=.95\linewidth]{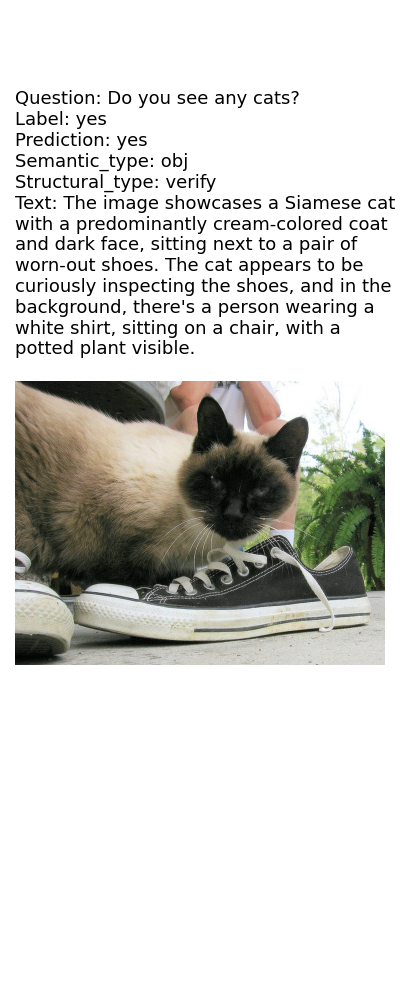}
  \label{fig:sfig11}
\end{subfigure}
\begin{subfigure}{.24\textwidth}
  \centering
  \includegraphics[trim= 0 210 0 0,width=.95\linewidth]{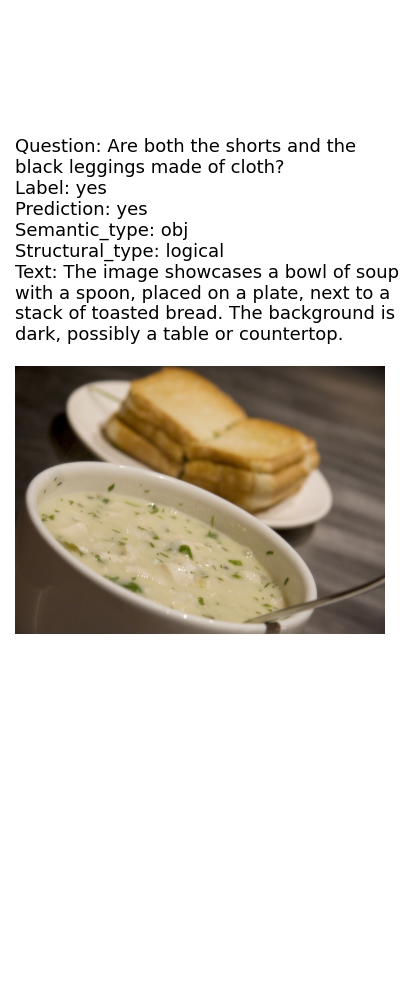}
  \label{fig:sfig12}
\end{subfigure}
\caption{Examples from \textbf{correct} predictions in case that QD image captioning is applied.}
\label{fig:fig}
\end{figure*}

\begin{figure*}
\begin{subfigure}{.24\textwidth}
  \includegraphics[trim= 0 210 0 0,clip,width=.95\linewidth]{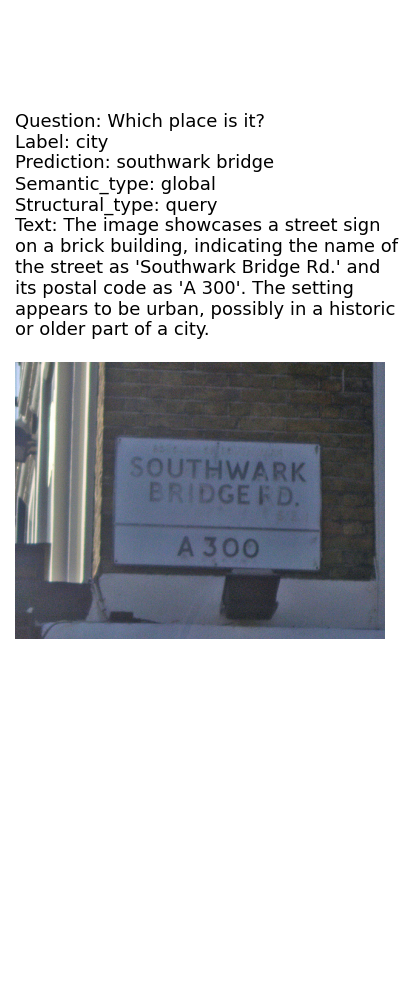}
  \label{fig:sfig1}
\end{subfigure}%
\begin{subfigure}{.24\textwidth}
  \includegraphics[trim= 0 210 0 0,clip,width=.95\linewidth]{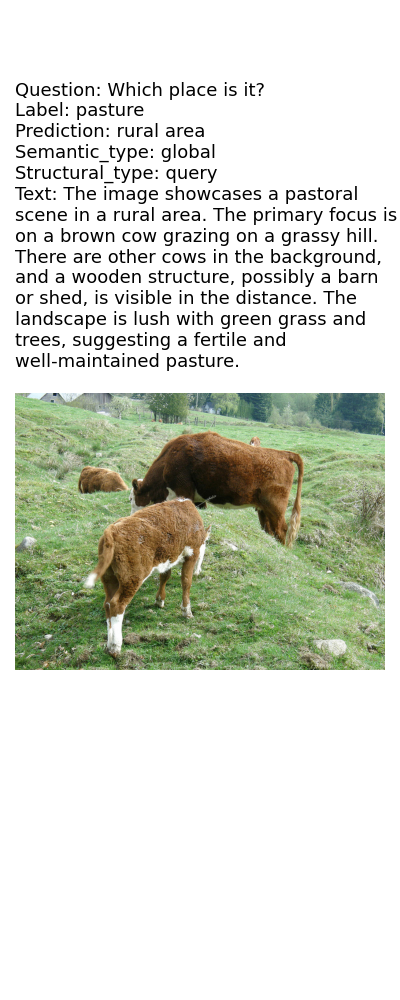}
  \label{fig:sfig2}
\end{subfigure}
\begin{subfigure}{.24\textwidth}
  \includegraphics[trim= 0 210 0 0,clip,width=.95\linewidth]{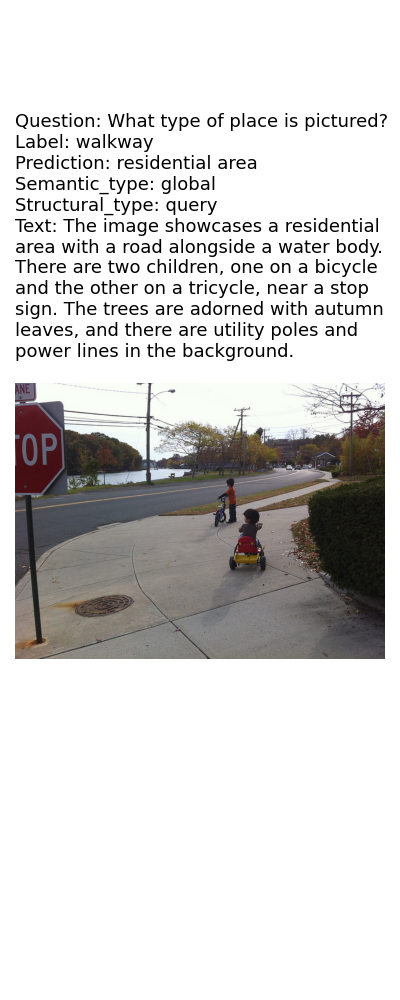}
  \label{fig:sfig3}
\end{subfigure}
\begin{subfigure}{.24\textwidth}
  \centering
  \includegraphics[trim= 0 210 0 0,clip,width=.95\linewidth]{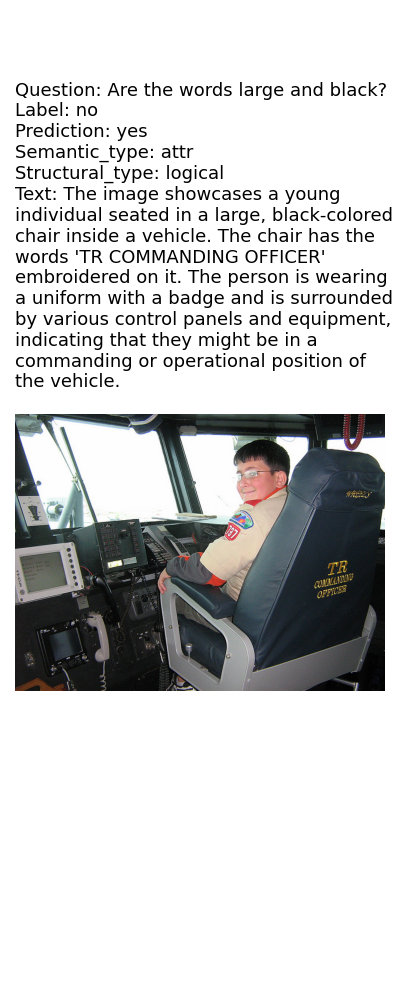}
  \label{fig:sfig4}
\end{subfigure}
\begin{subfigure}{.24\textwidth}
  \centering
  \includegraphics[trim= 0 210 0 0,width=.95\linewidth]{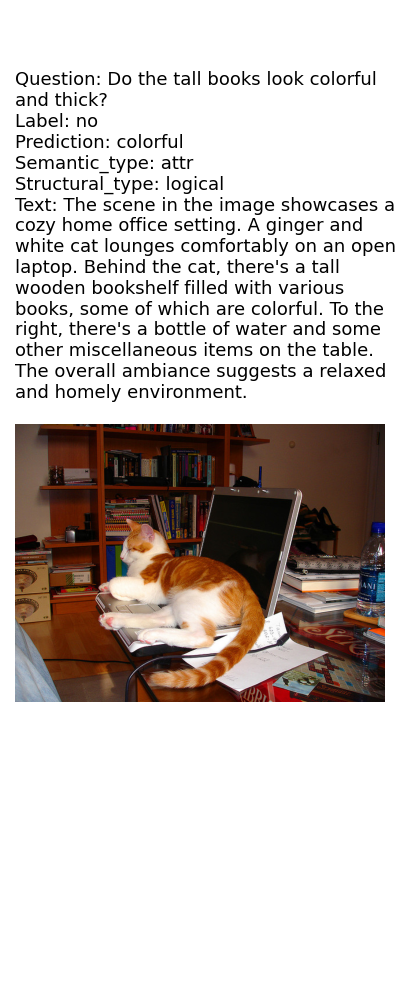}
  \label{fig:sfig5}
\end{subfigure}
\begin{subfigure}{.24\textwidth}
  \centering
  \includegraphics[trim= 0 210 0 0,width=.95\linewidth]{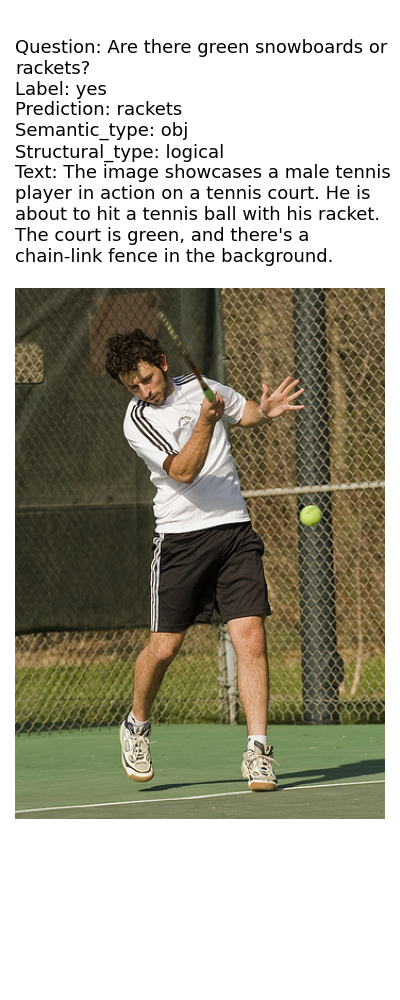}
  \label{fig:sfig6}
\end{subfigure}
\begin{subfigure}{.24\textwidth}
  \centering
  \includegraphics[trim= 0 210 0 0,width=.95\linewidth]{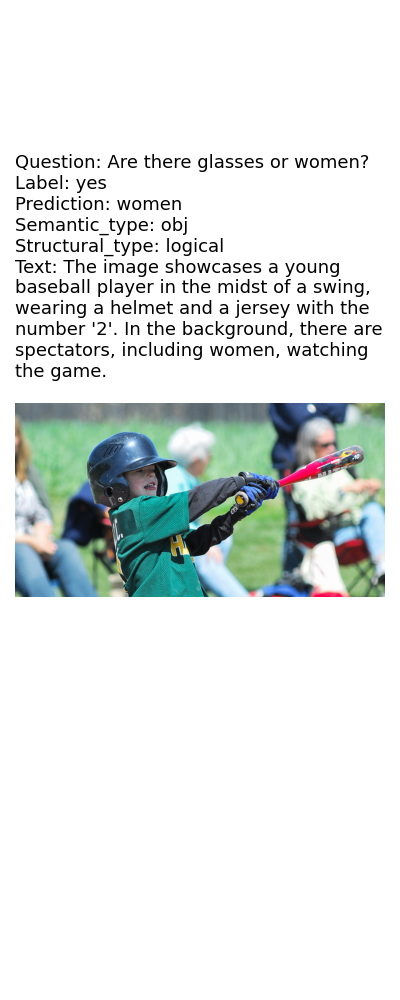}
  \label{fig:sfig7}
\end{subfigure}
\begin{subfigure}{.24\textwidth}
  \centering
  \includegraphics[trim= 0 210 0 0,width=.95\linewidth]{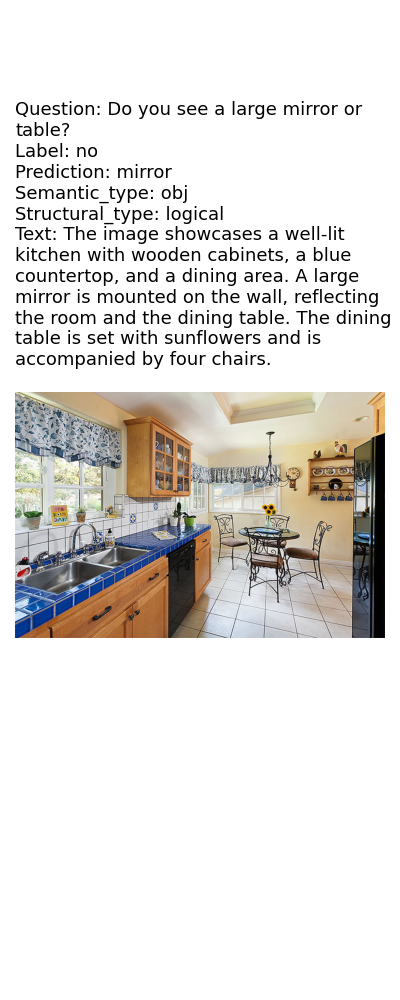}
  \label{fig:sfig8}
\end{subfigure}
\begin{subfigure}{.24\textwidth}
  \centering
  \includegraphics[trim= 0 210 0 0,width=.95\linewidth]{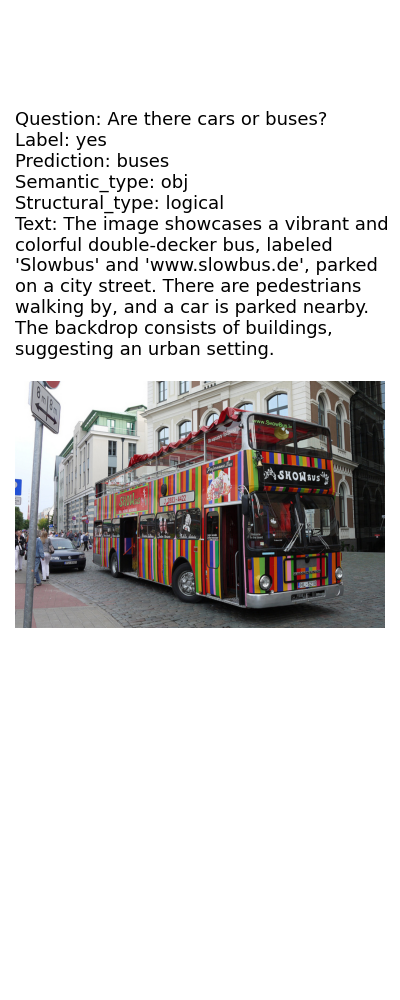}
  \label{fig:sfig9}
\end{subfigure}
\begin{subfigure}{.24\textwidth}
  \centering
  \includegraphics[trim= 0 210 0 0,width=.95\linewidth]{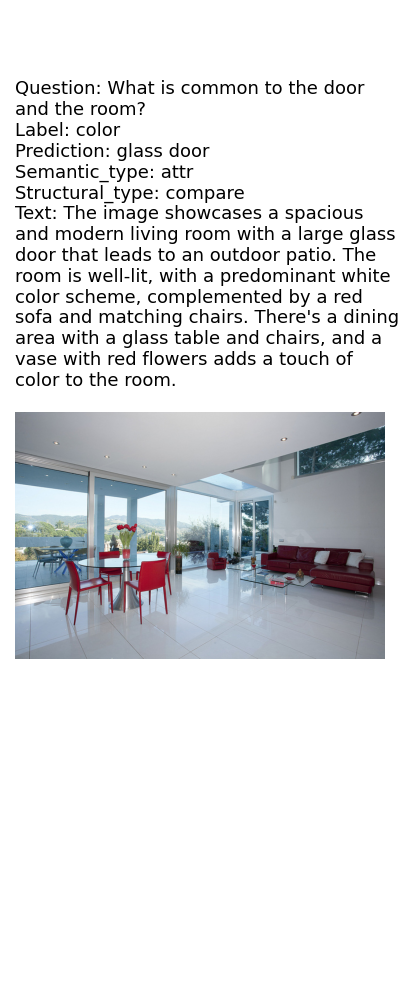}
  \label{fig:sfig10}
\end{subfigure}
\begin{subfigure}{.24\textwidth}
  \centering
  \includegraphics[trim= 0 210 0 0,width=.95\linewidth]{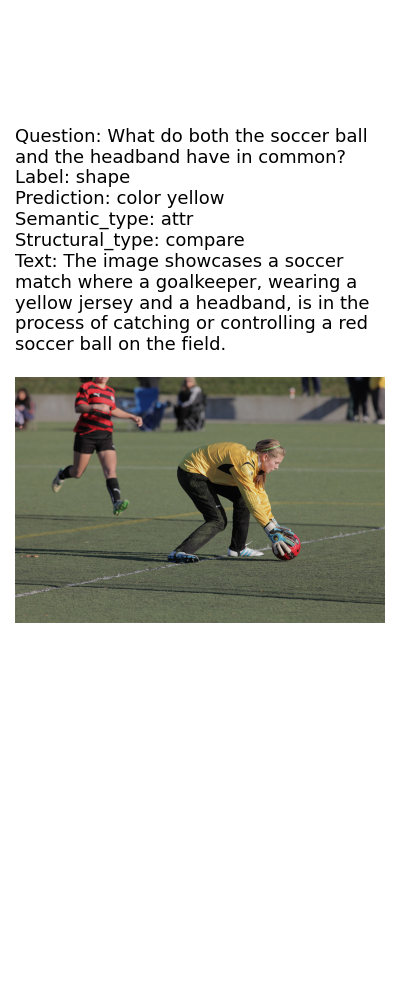}
  \label{fig:sfig11}
\end{subfigure}
\begin{subfigure}{.24\textwidth}
  \centering
  \includegraphics[trim= 0 210 0 0,width=.95\linewidth]{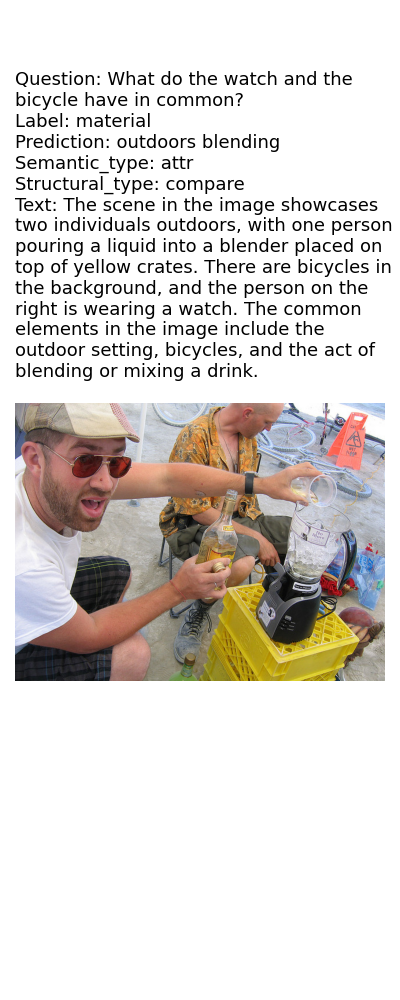}
  \label{fig:sfig12}
\end{subfigure}
\caption{Examples from \textbf{wrong} predictions in case that QD image captioning is applied.}
\label{fig:fig2}
\end{figure*}

\subsection{Main Findings}
\label{sec:main-findings}
Table \ref{tab:image-captioning} summarizes our results and demonstrates that employing our suggested QD image captioning approach for VQA enhances performance across most question categories compared to general-purpose image captioning. 
Also, Table \ref{tab:cosine-comp} indicates that the QD image captioning approach utilizing the CogVLM-chat variant surpasses other image captioning methods in evaluations seeking both different cosine similarity thresholds and exact matching.

\begin{table}[hbt!]
    \caption{Comparison of overall accuracy for exact matching (EM) and in different cosine similarity thresholds.}
    \centering
    \scalebox{0.80}{
    \begin{tabular}{llll} \toprule
    Models  & EM & sim=$0.9$ & sim=$0.8$ \\ \midrule
    CogVLM-C Cap. + GPT-3.5 QA & 36.77 & 38.21 & 43.01 \\
    CogVLM-V Cap. + GPT-3.5 QA & 36.21 & 37.51 & 41.21 \\ 
    CogVLM-C QD Cap. + GPT-3.5 QA & \textbf{37.64} & \textbf{39.24} & \textbf{44.48} \\ 
    CogVLM-C SB Cap. + GPT-3.5 QA & 34.14 & 35.06 & 39.41 \\ 
    FuseCap Cap. + GPT-3.5 QA & 33.17 & 34.18 & 37.64 \\
    BLIP-2 Cap. + GPT-3.5 QA & 34.77 & 35.53 & 39.11  \\ \midrule
    CogVLM VQA & \textbf{58.43} & \textbf{59.23} & \textbf{62.79}  \\ 
    BLIP-2 VQA & 37.82 & 38.57 & 42.33 \\ \bottomrule
    \end{tabular}}

    \label{tab:cosine-comp}
\end{table}

Significant performance enhancements are evident in QD image captions, particularly in the \emph{verify} category for yes/no questions, as well as \emph{attribute} and \emph{category} types primarily focused on identifying and describing a single object's properties.  However, challenges arise in the \emph{object} category often asking which of two objects exists in the frame. Particularly in this category of questions, despite the QD image captions containing relevant information, inaccuracies emerge due to the behavior of the QA model, as elaborated in Section \ref{sec:error-analysis}. 

We also notice that the QD captioning emphasizing question keywords is linked to a performance decline in the \emph{global} type questions. Global-type questions typically pertain to the overall content of an image. It suggests that the emphasis on question keywords in the caption negatively affects the model's ability to make inferences about the entire image. On the other hand, it is quite possible to give other answers to questions of this type that are meaningful and contextually correct but do not match the label. In most of the cases, we observe that GPT-3.5 predicts answers that could be correct but do not precisely match the expected label (see examples in Figure \ref{fig:fig2}).

In most question categories, the accuracy achieved by combining QD image captions with GPT-3.5 for VQA exceeds the performance of BLIP-2 FlanT5$_{XL}$ in the zero-shot setting.
However, all image captioning-based approaches indicate inferior performance compared to the CogVLM-chat model for VQA. We are intrigued to discover a notable disparity in performance when comparing the image captions extracted by the CogVLM-chat model and provided to LLM, in contrast to the VQA performance of the CogVLM-chat model, unlike the case with BLIP-2. 

Among the FuseCap, BLIP-2 OPT$_{2.7b}$, CogVLM-chat, and CogVLM-visual grounding models, the most informative captions for VQA are obtained through the CogVLM-chat variant. The CogVLM-visual grounding variant indicates the highest performance only in \emph{object} and \emph{logical} question categories. This suggests that visual grounding models may provide an advantage in these question categories with their capacity to connect language queries to relevant visual elements and reason about object-related relationships.

Limiting image captions to the most relevant sentence reduces the overall performance of the CogVLM-chat model, though the impact varies across question types, with \emph{verify}, \emph{query}, \emph{choose} and \emph{relation} types being more negatively affected. This suggests that sentences less directly related to the questions do not result in confusion or inaccuracies for LLM during the QA. Conversely, generating more comprehensive and context-rich image captions is necessary for optimal performance.

Figure \ref{fig:fig} and \ref{fig:fig2} feature examples of both correct and incorrect outcomes, where image captions are generated by the CogVLM-chat model using question-driven captioning and then fed to GPT-3.5 for answer prediction.

\subsection{Error Analysis}
\label{sec:error-analysis}
When examining incorrect predictions based on question types, we discover some common issues. 

We notice that 27\% of the incorrect predictions are related to yes/no questions.  A closer look reveals that in 11\% of the incorrectly answered yes/no questions, GPT-3.5 provides a response using a word other than \emph{yes} or \emph{no}.
For instance, when the provided caption is \emph{`The image showcases a skateboarder in action, possibly performing a trick on a ramp. The skateboarder is wearing protective gear, including a helmet, knee pads, and elbow pads. The background features a clear blue sky, trees, and a building. The overall ambiance suggests an outdoor skateboarding event or practice session.'}, in response to the question \emph{`Are there salt shakers or skateboards in the picture?'} GPT-3.5's prediction is \emph{skateboards}, while the ground-truth is \emph{yes}. 
We observe that most similar inaccuracies are associated with questions related to \emph{object} and \emph{logical} types, often connecting more than one object or attribute using conjunctions like \emph{and} or \emph{or}, as given in the example. We posit that this issue can be alleviated by crafting more effective prompts for GPT-3.5 or by employing a more powerful LLM for QA.

We also assess the instances where the LLM fails to provide an answer based on the information present in the image caption. Specifically, we examine the occurrences of \emph{not mentioned} and \emph{not visible} responses from GPT-3.5. Our findings indicate that, for the best general-purpose image captioning model, GPT-3.5 is not able to respond to 1.7\% of the questions. Notably, when employing question-driven captioning, this rate decreases to 0.5\%.

\section{Conclusion}
\label{sec:conclusion}
This study aims to develop a zero-shot VQA pipeline, leveraging LLMs with the inclusion of image captioning as an intermediate step, and evaluate its performance on the GQA benchmark. The proposed approach involves question-driven image captioning to transfer contextual information to the QA model. The study includes a thorough evaluation of zero-shot models for image captioning in the VQA context, comparing the impact of general-purpose and question-driven image captions in terms of various types of questions.
Our comparative analysis suggests that incorporating question-driven image captions into the VQA process has a more favorable effect on overall performance, surpassing the VQA performance of BLIP-2. Future endeavors may explore the integration of larger-scale LLMs, \eg, GPT-4, to further enhance performance. Additionally, evaluating the pipeline in a few-shot setting could offer a more comprehensive comparison. To enhance transparency, replacing the QA model with an interpretable alternative, such as graph-based QA models, can be explored.

\section*{Acknowledgements}
This work is partially supported by Middle East Technical University Scientific Research Projects Coordination Unit (METU-BAP), under the project number ADEP-704-2024-11482.


\end{document}